\documentclass[letterpaper,10 pt,conference]{ieeeconf}  % Comment this line out if you need a4paper

\usepackage{array}
\usepackage{times}
\usepackage{epsfig}
\usepackage{graphicx}
\usepackage{amsmath}
\usepackage{amssymb}
\usepackage{multirow}
\usepackage{setspace}
\usepackage{subfigure}
\usepackage{balance}
\newcolumntype{V}{>{$\vcenter\bgroup\hbox\bgroup}c<{\egroup\egroup$}}

\graphicspath{{../../../_under_review/2016_UAI1/imgs/},{../../../_under_review/2016_KAIS/figs-new/}}

\IEEEoverridecommandlockouts                              % This command is only needed if
\title{\LARGE \bf Identification of Promising Research Directions using \\Machine Learning Aided Medical Literature Analysis}

\author{Victor Andrei and Ognjen Arandjelovi\'c\\
School of Computer Science\\
University of St Andrews\\
St Andrews KY16 9SX\\
Fife, Scotland\\
United Kingdom}

\begin{document}

\maketitle
\thispagestyle{empty}
\pagestyle{empty}

%%%%%%%%%%%%%%%%%%%%%%%%%%%%%%%%%%%%%%%%%%%%%%%%%%%%%%%%%%%%%%%%%%%%%%%%%%%%%%%%
\begin{abstract}
  The rapidly expanding corpus of medical research literature presents major challenges in the understanding of previous work, the extraction of maximum information from collected data, and the identification of promising research directions. We present a case for the use of advanced machine learning techniques as an aide in this task and introduce a novel methodology that is shown to be capable of extracting meaningful information from large longitudinal corpora, and of tracking complex temporal changes within it.
\end{abstract}

%%%%%%%%%%%%%%%%%%%%%%%%%%%%%%%%%%%%%%%%%%%%%%%%%%%%%%%%%%%%%%%%%%%%%%%%%%%%%%%%
\section{Introduction}
Recent years have witnessed a remarkable convergence of two broad trends. The first of these concerns information i.e.\ data -- rapid technological advances coupled with an increased presence of computing in nearly every aspect of daily life, have for the first time made it possible to acquire and store massive amounts of highly diverse types of information. Concurrently and in no small part propelled by the environment just described, research in artificial intelligence -- in machine learning \cite{Aran2012g,Aran2012h,Aran2015,Aran2015d}, data mining~\cite{BeykAranPhunVenk+2014}, and pattern recognition, in particular -- has reached a sufficient level of methodological sophistication and maturity to process and analyse the collected data, with the aim of extracting novel and useful knowledge~\cite{Aran2015c,BeykAranPhunVenk+2014}. Though it is undeniably wise to refrain from overly ambitious predictions regarding the type of knowledge which may be discovered in this manner, at the very least it is true that few domains of application of the aforesaid techniques hold as much promise and potential as that of medicine and health in general.

Large amounts of highly heterogeneous data types are pervasive in medicine. Usually the concept of so-called ``big data'' in medicine is associated with the analysis of Electronic Health Records \cite{ChriElli2016,Aran2015g,Aran2016,VasiAran2016,VasiAran2016a}, large scale sociodemographic surveys of death causes \cite{RGI2009}, social media mining for health related data~\cite{BeykAranPhunVenk+2015} etc. Much less discussed and yet arguably no less important realm where the amount of information presents a challenge to the medical field is the medical literature corpus itself. Namely, considering the overarching and global importance of health (to say nothing of practical considerations such as the availability of funding), it is not surprising to observe that the amount of published medical research is immense and its growth is only continuing to accelerate. This presents a clear challenge to a researcher. Even restricted to a specified field of research, the amount of published data and findings makes it impossible for a human to survey the entirety of relevant publications exhaustively which inherently leads to the question as to what kind of important information or insight may go unnoticed or insufficiently appreciated. The premise of the present work is that advanced machine learning techniques can be used to assist a human in the analysis of this data. Specifically, we introduce a novel methodology based on Bayesian non-parametric inference that achieves this, as well as free software which researchers can use in the analysis of their corpora of interest.

\subsubsection{Previous work}
A limitation of most models described in the existing literature lies in their assumption that the data corpus is static. Here the term `static' is used to describe the lack of any associated temporal information associated with the documents in a corpus -- the documents are said to be exchangeable~\cite{BleiLaff2006a}. However, research articles are added to the literature corpus in a temporal manner and their ordering has significance. Consequently the topic structure of the corpus changes over time~\cite{Dyso2012,BeykPhunAranVenk2015,BeykAranPhunVenk2015a}: new ideas emerge, old ideas are refined, novel discoveries result in multiple ideas being related to one another thereby forming more complex concepts or a single idea multifurcating into different `sub-ideas' etc. The premise in the present work is that documents are not exchangeable at large temporal scales but can be considered to be at short time scales, thus allowing the corpus to be treated as \emph{temporally locally static}.

\section{Proposed approach\label{s:proposed}}
In this section we introduce our main technical contributions. We begin by reviewing the relevant theory underlying Bayesian mixture models, and then explain how the proposed framework employs these for the extraction of information from temporally varying document corpora.

\subsection{Bayesian mixture models}\label{ss:mixModels}
Mixture models are appropriate choices for the modelling of so-called heterogeneous data whereby heterogeneity is taken to mean that observable data is generated by more than one process (source). The key challenges lie in the lack of observability of the correspondence between specific data points and their sources, and the lack of \emph{a priori} information on the number of sources~\cite{RichGree1997}.

Bayesian non-parametric methods place priors on the infinite-dimensional space of probability distributions and provide an elegant solution to the aforementioned modelling problems. Dirichlet Process~(DP) in particular allows for the model to accommodate a potentially infinite number of mixture components~\cite{Ferg1973}:
\begin{align}
  p\left(x|\pi_{1:\infty},\phi_{1:\infty}\right)=\sum_{k=1}^{\infty}\pi_{k}f\left(x|\phi_{k}\right).
\end{align}
where $\text{DP}\left(\gamma,H\right)$ is defined as a distribution of
a random probability measure $G$ over a measurable space $\left(\Theta,\mathcal{B}\right)$,
such that for any finite measurable partition $\left(A_{1},A_{2},\ldots,A_{r}\right)$
of $\Theta$ the random vector $\left(G\left(A_{1}\right),\ldots,G\left(A_{r}\right)\right)$
is a Dirichlet distribution with parameters $\left(\gamma H\left(A_{1}\right),\ldots,\gamma H\left(A_{r}\right)\right)$. 

Owing to the discrete nature and infinite dimensionality of its draws, the DP is a useful prior for Bayesian mixture models. By associating different mixture components with atoms $\phi_{k}$, and assuming $x_{i}|\phi_{k}\overset{iid}{\sim}f\left(x_{i}|\phi_{k}\right)$ where $f\left(.\right)$ is the kernel of the mixing components, a Dirichlet process mixture model (DPM) is obtained~\cite{Radf2000}.

\subsubsection{Hierarchical DPMs}
While the DPM is suitable for the clustering of exchangeable data in a single group, many real-world problems are more appropriately modelled as comprising multiple groups of exchangeable data. In such cases it is desirable to model the observations of different groups jointly, allowing them to share their generative clusters. This so-called ``sharing of statistical strength'' emerges naturally when a hierarchical structure is implemented.

The DPM models each group of documents in a collection using an infinite number of topics. However, it is desired for multiple group-level DPMs to share their clusters. The hierarchical DP (HDP)~\cite{TehJordBealBlei2006} offers a solution whereby base measures of group-level DPs are drawn from a corpus-level DP. In this way the atoms of the corpus-level DP are shared across the documents; posterior inference is readily achieved using Gibbs sampling~\cite{TehJordBealBlei2006}.

\subsection{Modelling topic evolution over time\label{ss:contrib}}
We now show how the described HDP based model can be applied to the analysis of temporal topic changes in a \emph{longitudinal} data corpus.

Owing to the aforementioned assumption of a temporally locally static corpus we begin by discretizing time and dividing the corpus into epochs. Each epoch spans a certain contiguous time period and has associated with it all documents with timestamps within this period. Each epoch is then modelled separately using a HDP, with models corresponding to different epochs sharing their hyperparameters and the corpus-level base measure. Hence if $n$ is the number of epochs, we obtain $n$ sets of topics
$\boldsymbol{\phi}=\left\{ \boldsymbol{\phi}_{t_{1}},\ldots,\boldsymbol{\phi}_{t_{n}}\right\} $
where $\boldsymbol{\phi}_{t}=\left\{ \theta_{1,t},\ldots,\phi_{K_{t},t}\right\} $
is the set of topics that describe epoch $t$, and $K_{t}$ their number. 

\subsubsection{Topic relatedness\label{ss:similarity}}
Our goal now is to track changes in the topical structure of a data corpus over time. The simplest changes of interest include the emergence of new topics, and the disappearance of others. More subtly, we are also interested in how a specific topic changes, that is, how it evolves over time in terms of the contributions of different words it comprises. Lastly, our aim is to be able to extract and model complex structural changes of the underlying topic content which result from the interaction of topics. Specifically, topics, which can be thought of as collections of memes, can merge to form new topics or indeed split into more nuanced memetic collections. This information can provide valuable insight into the refinement of ideas and findings in the scientific community, effected by new research and accumulating evidence.

The key idea behind our tracking of simple topic evolution stems from the observation that while topics may change significantly over time, changes between successive epochs are limited. Therefore we infer the continuity of a topic in one epoch by relating it to all topics in the immediately subsequent epoch which are sufficiently similar to it under a suitable similarity measure -- we adopt the well known Bhattacharyya distance (BHD). This can be seen to lead naturally to a similarity graph representation whose nodes correspond to topics and whose edges link those topics in two epochs which are related. Formally, the weight of the directed edge that links $\phi_{j,t}$, the $j$-th topic in epoch $t$, and $\phi_{k,t+1}$ is $\rho_\text{BHD}\left(\phi_{j,t},\phi_{k,t+1}\right)$ where $\rho_\text{BHD}$ denotes the BHD.

In constructing a similarity graph a threshold to used to eliminate automatically weak edges, retaining only the connections between sufficiently similar topics in adjacent epochs. Then the disappearance of a particular topic, the emergence of new topics, and gradual topic evolution can be determined from the structure of the graph. In particular if a node does not have any edges incident to it, the corresponding topic is taken as having emerged in the associated epoch. Similarly if no edges originate from a node, the corresponding topic is taken to vanish in the associated epoch. Lastly when exactly one edge originates from a node in one epoch and it is the only edge incident to a node in the following epoch, the topic is understood as having evolved in the sense that its memetic content may have changed.

\begin{figure*}[t]
  \centering
  \subfigure[Topic speciation]{\includegraphics[width=0.8\columnwidth]{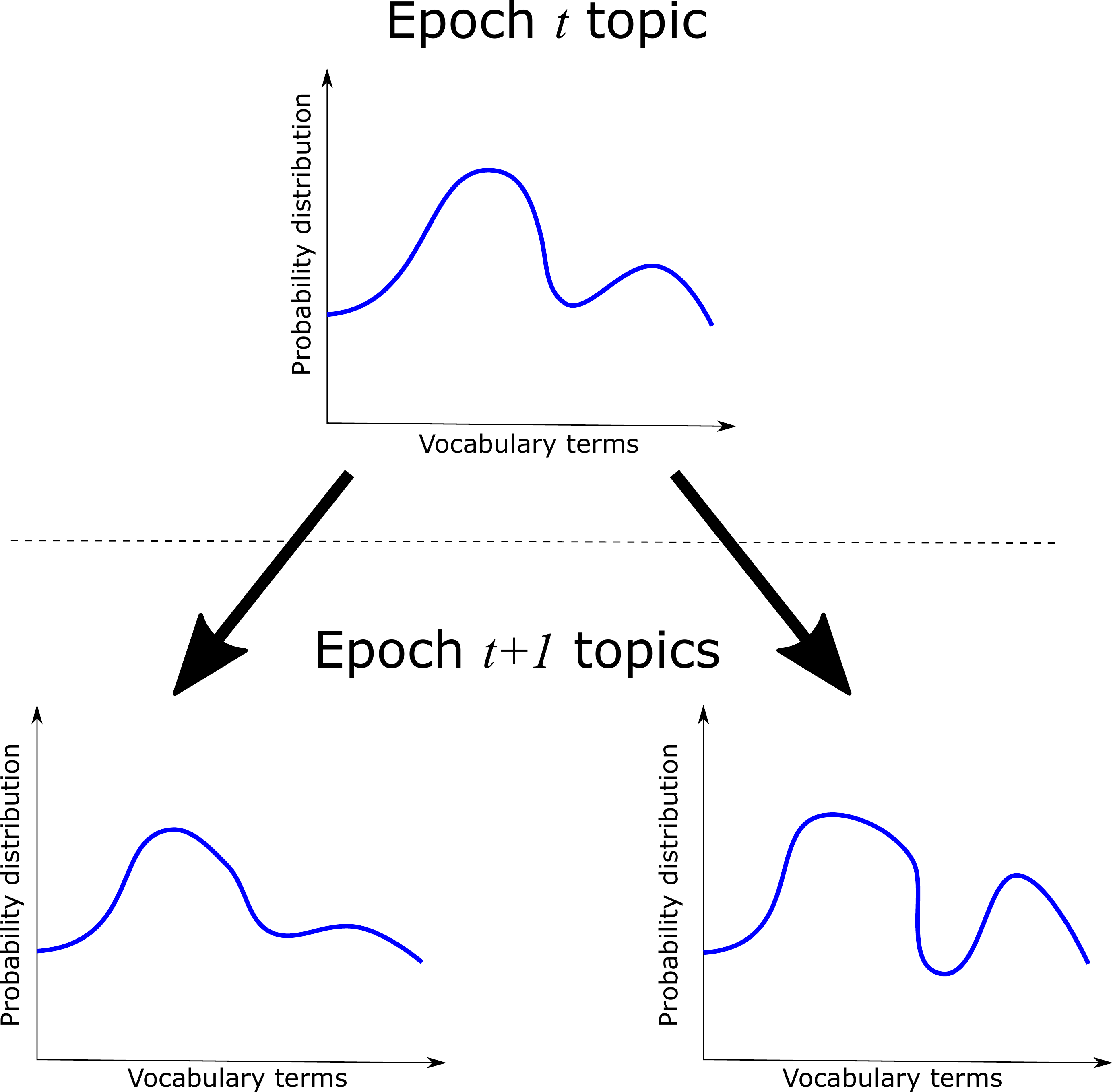}\label{f:speciation}}~~~~~~~~~~~~~~~~~~~~~~~
  \subfigure[Topic splitting]{\includegraphics[width=0.8\columnwidth]{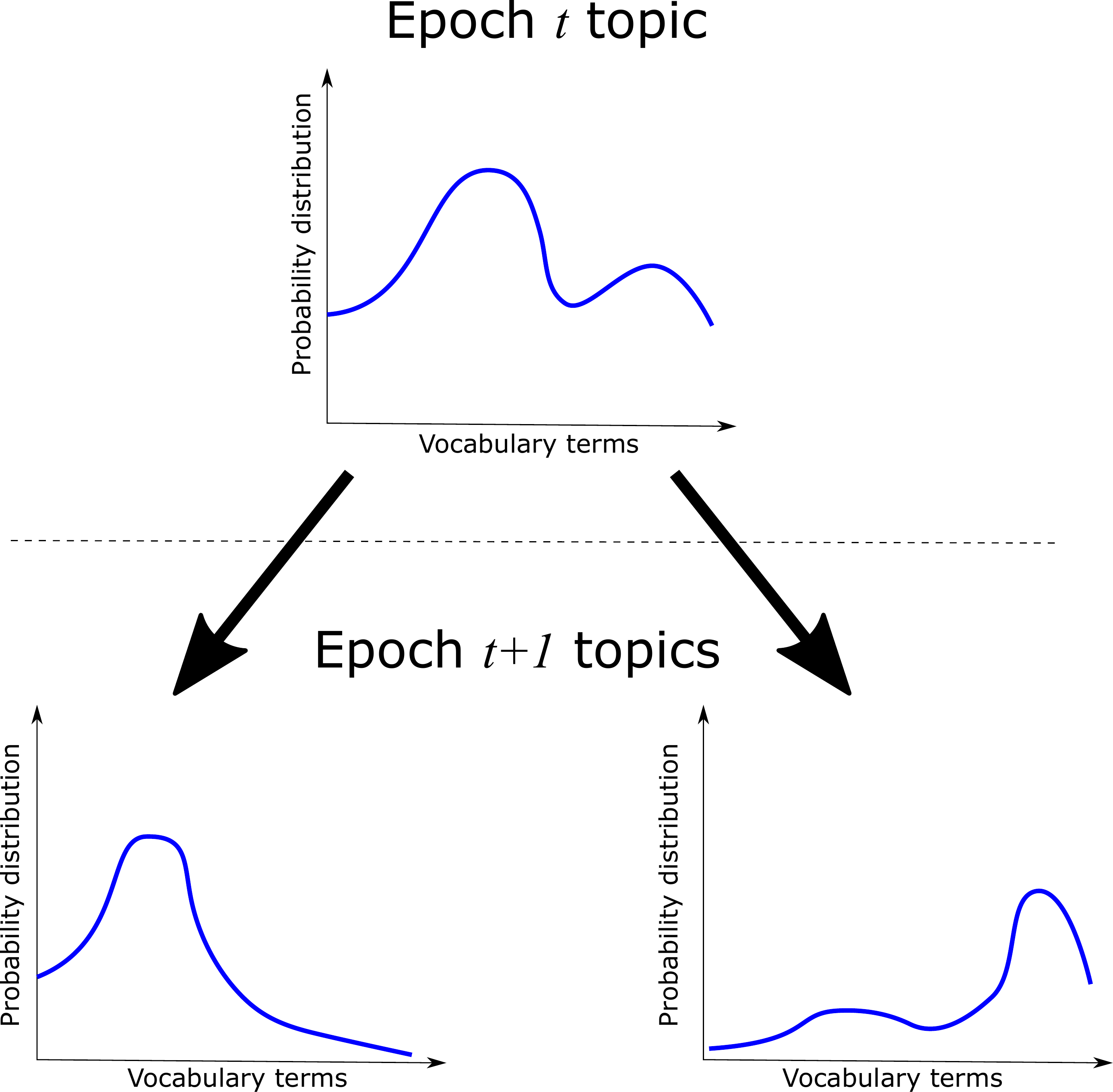}\label{f:splitting}}
  \caption{ This paper is the first work to describe the difference between two topic evolution phenomena: (a) topic speciation and (b) topic splitting. }
\end{figure*}

A major challenge to the existing methods in the literature concerns the detection of topic merging and splitting. Since the connectedness of topics across epochs is based on their similarity what previous work describes as `splitting' or indeed `merging' does not adequately capture these phenomena. Rather, adopting the terminology from biological evolution, a more accurate description would be `speciation' and `convergence' respectively. The former is illustrated in Fig~\ref{f:speciation} whereas the latter is entirely analogous with the time arrow reversed. What the conceptual diagram shown illustrates is a slow differentiation of two topics which originate from the same `parent'. Actual topic splitting, which does not have a biological equivalent in evolution, and which is conceptually illustrated in Fig~\ref{f:splitting} cannot be inferred by measuring topic similarity. Instead, in this work we propose to employ the Kullback-Leibler divergence (KLD) for this purpose. This divergence is asymmetric can be intuitively interpreted as measuring how well one probability distribution `envelops' another. KLD between two probability distributions $p(i)$ and $q(i)$ is defined as follows:
\begin{align}
  \rho_\text{KLD} = \sum_i p(i) \log \frac{p(i)}{q(i)}
\end{align}
It can be seen that a high penalty is incurred when $p(i)$ is significant and $q(i)$ is low. Hence, we use the BHD to track gradual topic evolution, speciation, and convergence, while the KLD (computed both in forward and backward directions) is used to detect topic splitting and merging.

\subsubsection{Automatic temporal relatedness graph construction\label{ss:construction}}
Another novelty of the work first described in this paper concerns the building of the temporal relatedness graph. We achieve this almost entirely automatically, requiring only one free parameter to be set by the user. Moreover the meaning of the parameter is readily interpretable and understood by a non-expert, making our approach highly usable.

Our methodology comprises two stages. Firstly we consider all inter-topic connections present in the initial fully connected graph and extract the empirical estimate of the corresponding cumulative density function (CDF). Then we prune the graph based on the operating point on the relevant CDF. In other words if $F_\rho$ is the CDF corresponding to a specific initial, fully connected graph formed using a particular similarity measure (BHD or KLD), and $\zeta \in [0, 1]$ the CDF operating point, we prune the edge between topics $\phi_{j,t}$ and $\phi_{k,t+1}$ iff $\rho(\phi_{j,t},\phi_{k,t+1}) < F^{-1}_\rho (\zeta)$.

\section{Evaluation and discussion}
We now analyse the performance of the proposed framework empirically on a large real world data set. 

\subsection{Evaluation data}\label{sss:rawData}
We used the PubMed interface to access the US National Library of Medicine and retrieve from it scholarly articles. We searched for publication on the metabolic syndrome (MetS) using the keyphrase``metabolic syndrome''  and collected papers written in English. The earliest publication found was that by Berardinelli~\textit{et al.}~\cite{BeraCorddeAlCouc1953}. We collected all matching publications up to the final one indexed by PubMed on 10th Jan 2016, yielding a corpus of 31,706 publications. %We used their abstracts to evaluate our method.

\subsubsection{Pre-processing\label{sss:preprocessing}}
The raw data collected from PubMed is in the form of free text. To prepare it for automatic analysis a series of `pre-processing' steps are required. The goal is to remove words which are largely uninformative, reduce dispersal of semantically equivalent terms, and thereafter select terms which are included in the vocabulary over which topics are learnt.

We firstly applied soft lemmatization using the WordNet$^\circledR$ lexicon~\cite{Mill1995} to normalize for word inflections. No stemming was performed to avoid semantic distortion often effected by heuristic rules used by stemming algorithms. After lemmatization and the removal of so-called stop-words, we obtained approximately 3.8 million terms in the entire corpus when repetitions are counted, and 46,114 unique terms. Constructing the vocabulary for our method by selecting the most frequent terms which explain 90\% of the energy in a specific corpus resulted in a vocabulary containing 2,839 terms.

\subsection{Results}
\begin{figure}
  \centering
  \subfigure[BHD-KLD normalized overlap]{\includegraphics[width=0.99\columnwidth]{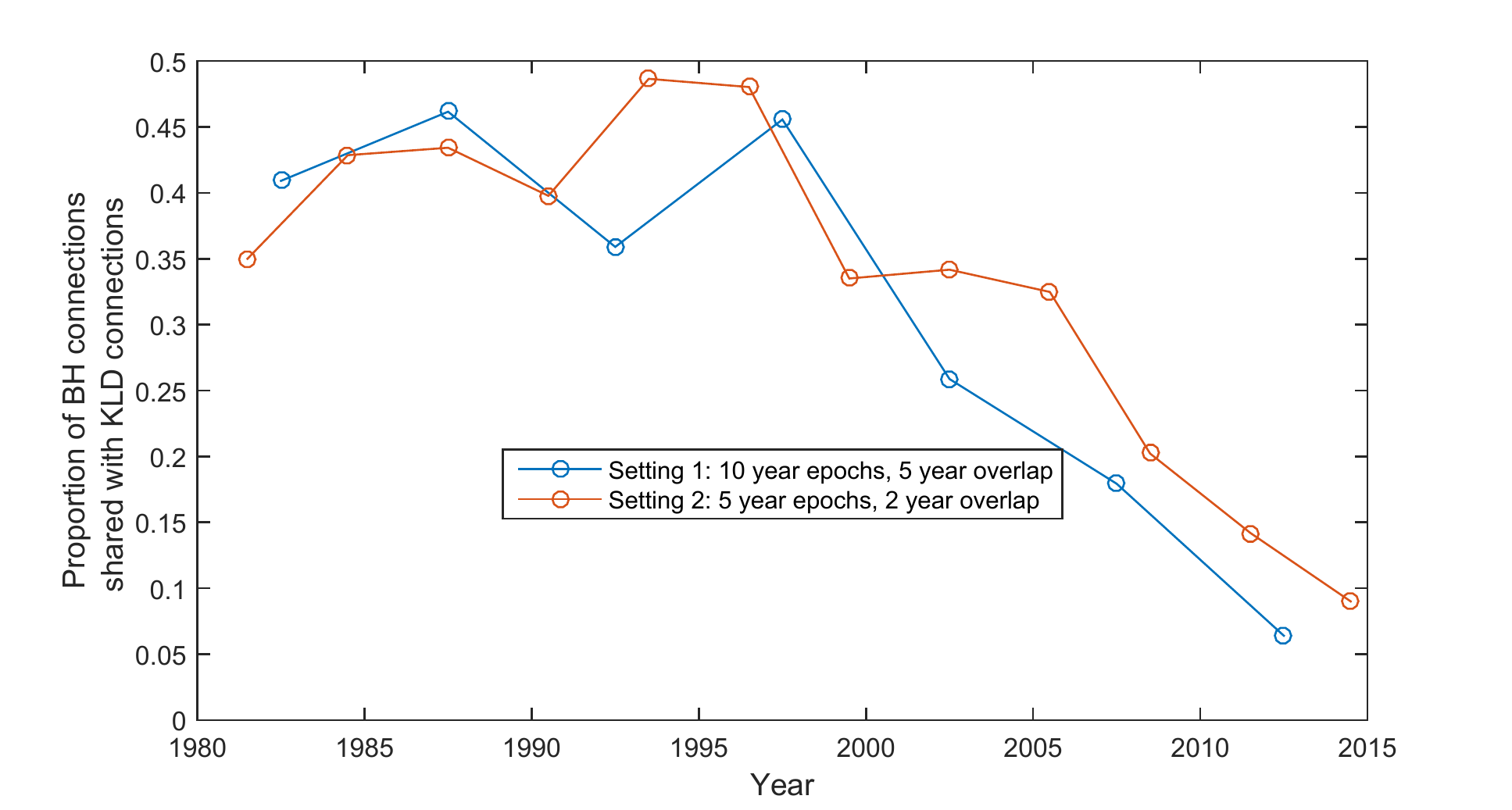}}
  \subfigure[KLD-BHD normalized overlap]{\includegraphics[width=0.99\columnwidth]{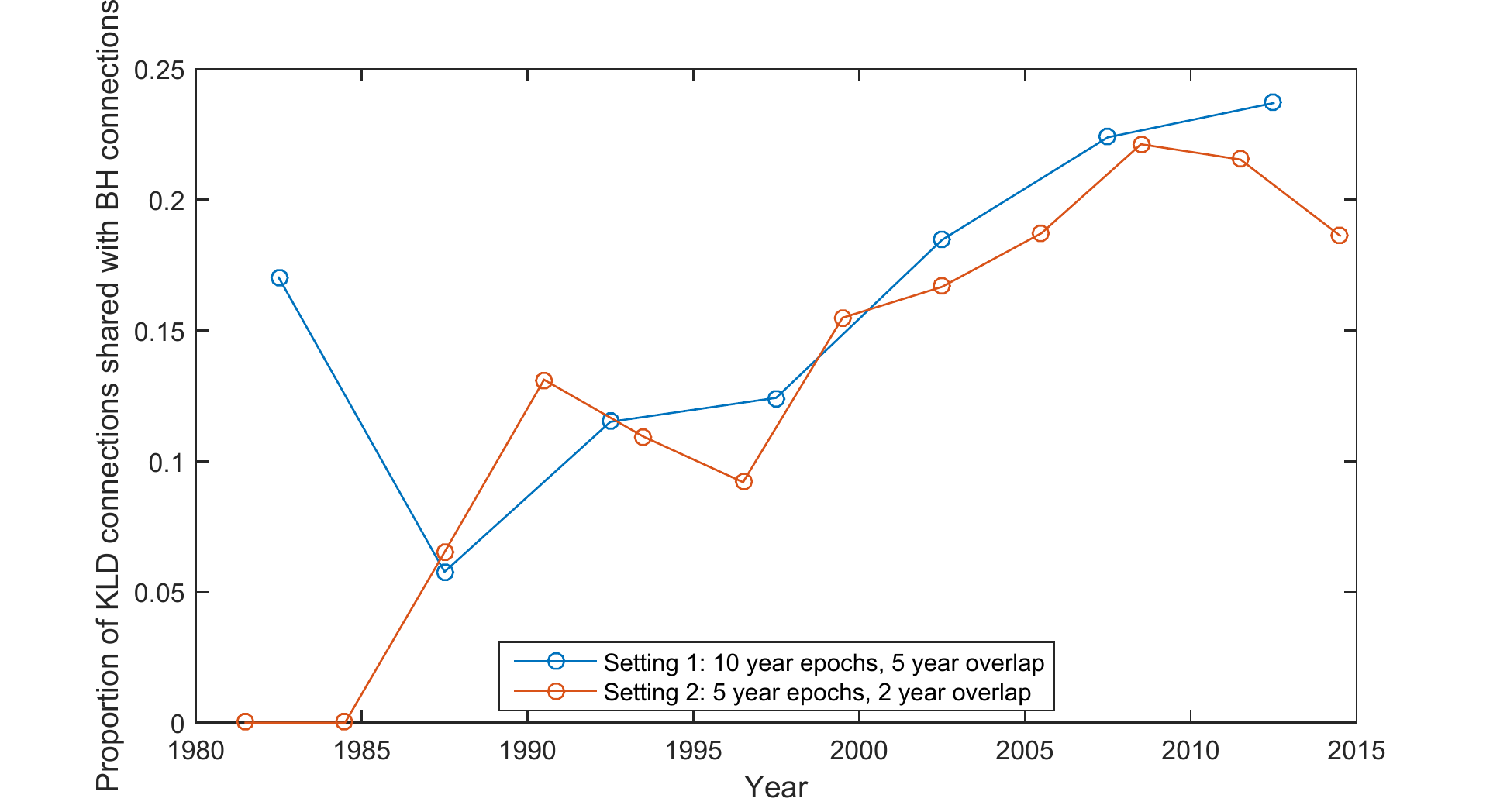}}
  \caption{ The proportion of topic connections shared between the BHD and the KLD temporal relatedness graphs, normalized by (a) the number of BHD connections, and (b) the number of KLD connections, in an epoch. }
  \label{f:common}
\end{figure}

We stared evaluation by examining whether the two topic relatedness measures (BHD and KLD) are capturing different aspects of relatedness. To obtain a quantitative measure we looked at the number of inter-topic connections formed in respective graphs both when the BHD is used as well as when the KLD is applied instead. The results were normalized by the total number of connections formed between two epochs, to account for changes in the total number of topics across time. Our results are summarized in Fig~\ref{f:common}. A significant difference between the two graphs is readily evident -- across the entire timespan of the data corpus, the number of Bhattacharyya distance based connections also formed through the use of the KLD is less than 40\% and in most cases less than 30\%. An even greater difference is seen when the proportion of the KLD connections is examined -- it is always less than 25\% and most of the time less than 15\%.

\begin{figure}
  \centering
  \includegraphics[width=0.99\columnwidth]{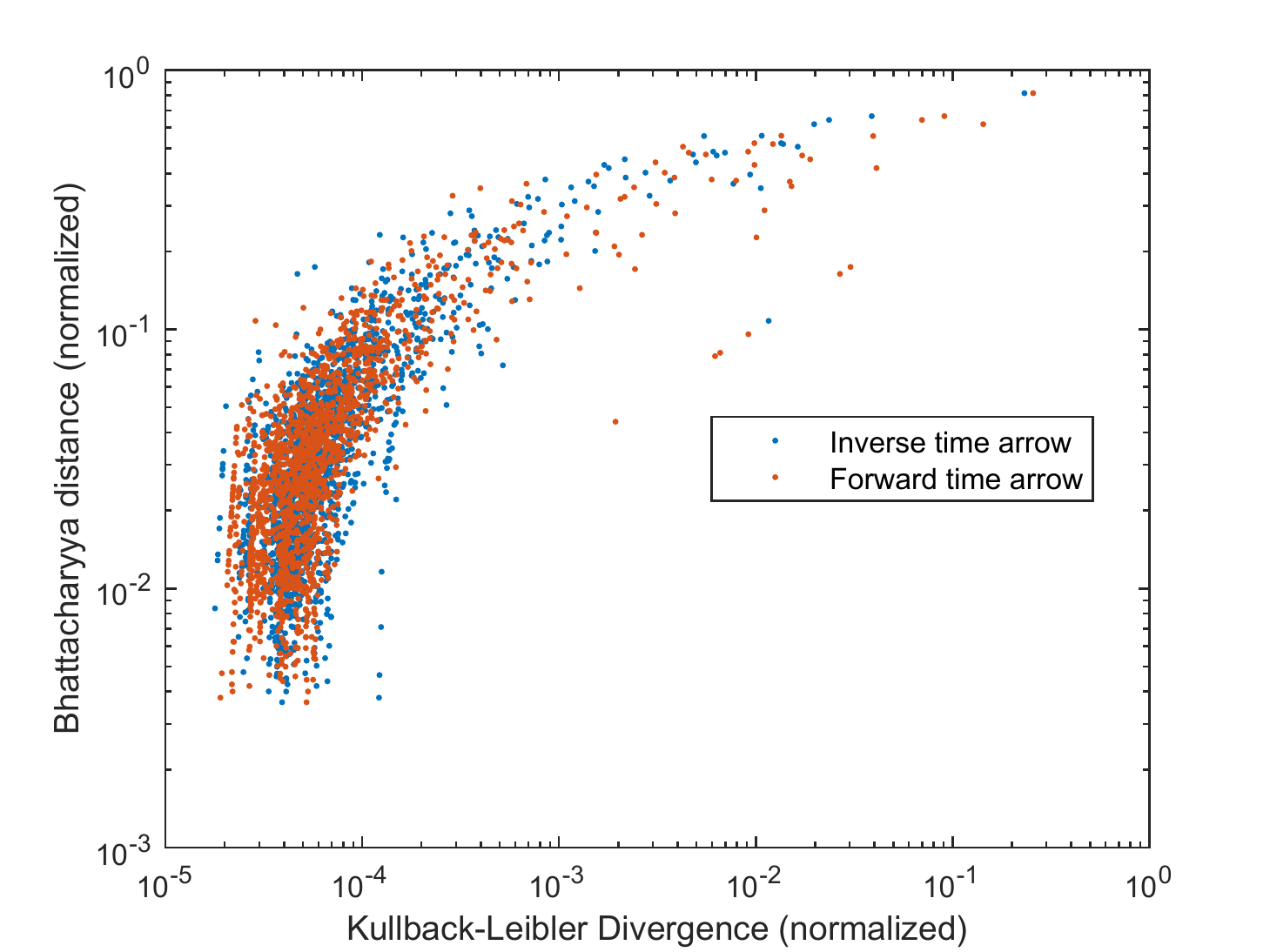}
  \caption{ Relationship between inter-topic edge strengths computed using the BHD and the KLD before the pruning of the respective graphs. }
  \label{f:smooth}
\end{figure}

To get an even deeper insight into the contribution of the two relatedness measures, we examined the corresponding topic graphs before edge pruning. The plot in Fig~\ref{f:smooth} shows the variation in inter-topic edge strengths computed using the BHD and the KLD (in forward and backward directions) -- the former as the $x$ coordinate of a point corresponding to a pair of topics, and the latter as its $y$ coordinate. The scatter of data in the plot corroborates our previous observation that the two similarity measures indeed do capture different aspects of topic behaviour.

We performed extensive qualitative analysis which is necessitated by the nature of the problem at hand and the so-called `semantic gap' that underlies it. In all cases we found that our algorithm revealed meaningful and useful information, as confirmed by an expert in the area of metabolic MetS research.

Our final contribution comprises a web application which allows users to upload and analyse their data sets using the proposed framework. The application allows a range of powerful tasks to be performed quickly and in an intuitive manner. For example, the user can search for a given topic using keywords (and obtain a ranked list), trace the origin of a specific topic backwards in time, or follow its development in the forward direction, examine word clouds associated with topics, display a range of statistical analyses, or navigate the temporal relatedness graph.

\section{Summary and Conclusions}
In this work we presented a case for the importance of use of advanced machine learning techniques in the analysis and interpretation of medical literature. We described a novel framework based on non-parametric Bayesian techniques which is able to extract and track complex, semantically meaningful changes to the topic structure of a longitudinal document corpus. Moreover this work is the first to describe and present a method for differentiating between two types of topic structure changes, namely topic splitting and what we termed topic speciation. Experiments on a large corpus of medical literature concerned with the metabolic syndrome was used to illustrate the performance of our method. Lastly, we developed a web application which allows users such as medical researchers to upload their data sets and apply our method for their analysis; the application and its code will be made freely available following publication.

\balance

\bibliographystyle{ieee}
\bibliography{../../../my_bibliography,../../../oa_physiology}

\end{document}